\pdfoutput=1

\documentclass[11pt,a4paper]{article}
\usepackage[utf8]{inputenc}
\usepackage[T1]{fontenc}
\usepackage[margin=2.5cm]{geometry} 
\usepackage{amssymb, amsmath, float, bm, url}
\usepackage{graphicx}
\usepackage{caption}
\usepackage{subcaption}
\usepackage{makecell}

\usepackage[numbers,sort&compress]{natbib}

\DeclareRobustCommand{\erase}{\bgroup\markoverwith{\textcolor{red}{\rule[.5ex]{2pt}{0.4pt}}}\ULon}

\title{Estimation of geometric transformation matrices using grid-shaped pilot signals
\thanks{This is the author's accepted manuscript of the paper to be published in APSIPA Transactions on Signal and Information Processing (DOI: 10.1561/116.20250067).}}

\author{Rinka Kawano, Masaki Kawamura \\
Graduate School of Sciences and Technology for Innovation, Yamaguchi University}

\begin{document}
\maketitle

\begin{abstract}
Digital watermarking techniques are essential to prevent unauthorized use of images. Since pirated images are often geometrically distorted by operations such as scaling and cropping, accurate synchronization - detecting the embedding position of the watermark - is critical for proper extraction. In particular, cropping changes the origin of the image, making synchronization difficult. However, few existing methods are robust against cropping. To address this issue, we propose a watermarking method that estimates geometric transformations applied to a stego image using a pilot signal, allowing synchronization even after cropping. A grid-shaped pilot signal with distinct horizontal and vertical values is embedded in the image. When the image is transformed, the grid is also distorted. By analyzing this distortion, the transformation matrix can be estimated. Applying the Radon transform to the distorted image allows estimation of the grid angles and intervals. In addition, since the horizontal and vertical grid lines are encoded differently, the grid orientation can be determined, which reduces ambiguity. To validate our method, we performed simulations with anisotropic scaling, rotation, shearing, and cropping. The results show that the proposed method accurately estimates transformation matrices with low error under both single and composite attacks.
\end{abstract}

\vspace{10pt}
\vspace{20pt}

\section{Introduction}

Digital watermarking, a technology for protecting digital content, is attracting attention as the problem of image piracy and unauthorized use of images becomes more serious. Digital watermarking~\cite{hiding} is a technique for covertly embedding other information in digital content, such as still images, video, and audio.
The information to be embedded is called a watermark, and an image with a watermark embedded is called a stego image. In multibit watermarking, the information to be embedded is referred to as the message and the encoded information is referred to as the watermark in order to distinguish between them. By pre-embedding the owner's ID or signature in images to be posted on social media, the content can be controlled. For example, by detecting a watermark in misused content, it is possible to claim true ownership of the content. However, many pirated images are manipulated by scaling or cropping.
Additionally, images are compressed when saved. These attacks on stego images, such as image manipulation and compression, can make the watermark disappear or difficult to detect. Therefore, robust watermarking methods against various attacks need to be considered.

Attacks on images are divided into non-geometric and geometric attacks. Non-geometric attacks are those that change pixel values, such as JPEG compression and noise addition. When a stego image is degraded by a non-geometric attack, the embedded watermark is degraded. However, the coordinates where the watermark is embedded remain unchanged. Therefore, the watermark can be extracted from its original location and the exact message can be decoded if the message is encoded using error-correcting codes~\cite{BCH} or spread spectrum techniques~\cite{cox_spectrum,spread_spectrum}.
Many conventional methods are robust against non-geometric attacks by introducing these techniques.
Geometric attacks are those that change the position of pixels, such as scaling, rotating, or cropping an image.
When a stego image is degraded by a geometric attack, the coordinates of the embedded watermark are lost. Therefore, it is necessary to detect the position of the watermark, i.e., to achieve synchronization.
Watermarking using SIFT features~\cite{sift_watermark} is one of the most robust methods against geometric attacks. By embedding a watermark around multiple SIFT feature points, resistance against geometric attacks can be achieved~\cite{hayashi,circularly_DFT,localized_DFT}. However, the detection performance is still insufficient.

Many conventional watermarking methods that claim robustness against geometric attacks assume the use of clipped images for detection. It is important to distinguish between {\it clipping} and {\it cropping}~\cite{lin2001rotation}: clipping refers to the process of padding certain regions of an image with zeros, whereby the position of the watermark remains known because the location of the clipped region is predetermined. In contrast, cropping involves the complete removal of image regions, making it difficult to determine the location of the embedded watermark. 
For example, Lin {\it et al.}~\cite{lin2001rotation} proposed an RST-resistant method that embeds a watermark in the amplitude of the log-polar coordinates of the Fourier transform. This method is resistant to rotation, scaling, and translation. However, it is not resistant to cropping.
In the methods of Pereira \cite{pereira2000robust} and Pun and Kang {\it et al.}~\cite{kang2003dwt}, an original image is padded to a fixed size and then Fourier transformed, embedding both a template for detecting geometric transformations and a watermark in the coefficients. Similarly, the image is padded to a fixed size and inversely transformed based on the transformation matrix obtained from the template when the watermark is extracted from a geometrically transformed image. It is claimed that these operations enable accurate detection of the watermark even from images that have undergone geometric attacks and cropping.
In addition, in the method of Hu and Xiang \cite{hu2020cover}, a watermark is embedded in the coefficients of the Zernike moments. Zernike moments not only have the property of being invariant with respect to rotation, but they also become robust to scaling by normalizing their amplitudes.
However, a brute force search is required to synchronize the images when using Fourier transforms or Zernike moments if the origin of the attacked image changes due to cropping. 
Brute-force algorithms are often employed to achieve synchronization, but such methods are computationally expensive. Consequently, watermarking techniques that rely on brute-force synchronization cannot be regarded as truly robust against geometric attacks.
Thus, effective estimation methods for geometric attacks involving cropping have yet to be established. Section 7 of the survey paper~\cite{wan2022comprehensive} also states that the problem of geometric attacks involving cropping is still a difficult problem.
If the type and strength of the geometric attack can be estimated, the position of the watermark can be accurately determined, thereby reducing the number of detection errors. Based on this idea, we focus on the communication channel estimation framework.
Communication channel estimation schemes often employ a technique that uses a pilot signal that is distinct from the actual message. This pilot signal is degraded as it passes through a noisy communication channel, and the channel parameters can be estimated based on this degradation.
In our watermarking scheme, we introduced the concept of a pilot signal and proposed a method to embed it into an image~\cite{IEICE2024,apsipa2024}. When the stego image is subjected to an attack, the pilot signal is also degraded. By analyzing the distortion of the pilot signal, the parameters of the attack can thus be estimated.

Su {\it et al.}~\cite{geometrically} proposed a watermarking method that incorporates pilot signals. In their approach, the watermark is embedded in regions surrounding SIFT feature points, and pilot signals are used as reference points to identify these embedding regions. Accordingly, the watermark region can be located by detecting the feature points associated with the embedded pilot signals.
However, their method uses the pilot signal only for synchronization purposes and not for estimating the parameters of geometric attacks.

We focus on the fact that pilot signals can be used not only for synchronization, but also for estimating the parameters of geometric attacks.
To enable the estimation of geometric transformations applied to an image using pilot signals, it is essential to design an appropriate signal shape and select an embedding domain that allows the effects of such attacks to be accurately detected.
In our previous work \cite{apsipa2024}, we proposed a method for embedding a pilot signal in the form of a grid. During detection, the extracted pilot signal was projected in the vertical and horizontal directions, and the slopes and spacings of the grid lines were measured from the histograms of signal intensity. This approach enabled the estimation of geometric attacks, such as scaling and rotation.
In the proposed method, the grid-shaped signal is embedded in the image as a pilot signal. When this signal is transformed using the Radon transform, the slopes and intervals of the grid lines become measurable. Based on these slopes and intervals, the transformation matrix corresponding to the geometric attack can be estimated.
Furthermore, by embedding pilot signals with different values in the horizontal and vertical directions, the orientation of the orthogonal grid lines can be uniquely determined.
In this paper, we propose a method for detecting a degraded pilot signal and estimating the geometric transformation matrix from a geometrically attacked stego image. The effectiveness of the proposed method is demonstrated by computer simulations, which confirm the accurate estimation of the transformation matrix.
We demonstrate that it is possible to estimate individual transformations, including shear, scaling, and rotation transformations, as well as their composite transformations.

This paper is organized as follows:
Chapter~\ref{sec:proposed_method} provides an overview of the proposed method.
Chapter~\ref{sec:EstimationSimulation} details the performance evaluation of attack estimation. Chapter~\ref{sec:WatermarkSimulation} presents the performance evaluation of applying the method to a watermarking method.
Chapter~\ref{sec:conclusion} concludes the paper.

\section{Proposed method} \label{sec:proposed_method}
In the proposed method, the geometric attack applied to the stego image can be estimated using a grid-shaped pilot signal. Once this signal is distorted by the geometric attack, the slope and interval of the grid lines change. These changes can be detected by Radon transform of the signal.
\subsection{Procedure for embedding a pilot signal} \label{subsec:embed}

\begin{figure}[tb]
  \centering
  \includegraphics[width=0.6\linewidth]{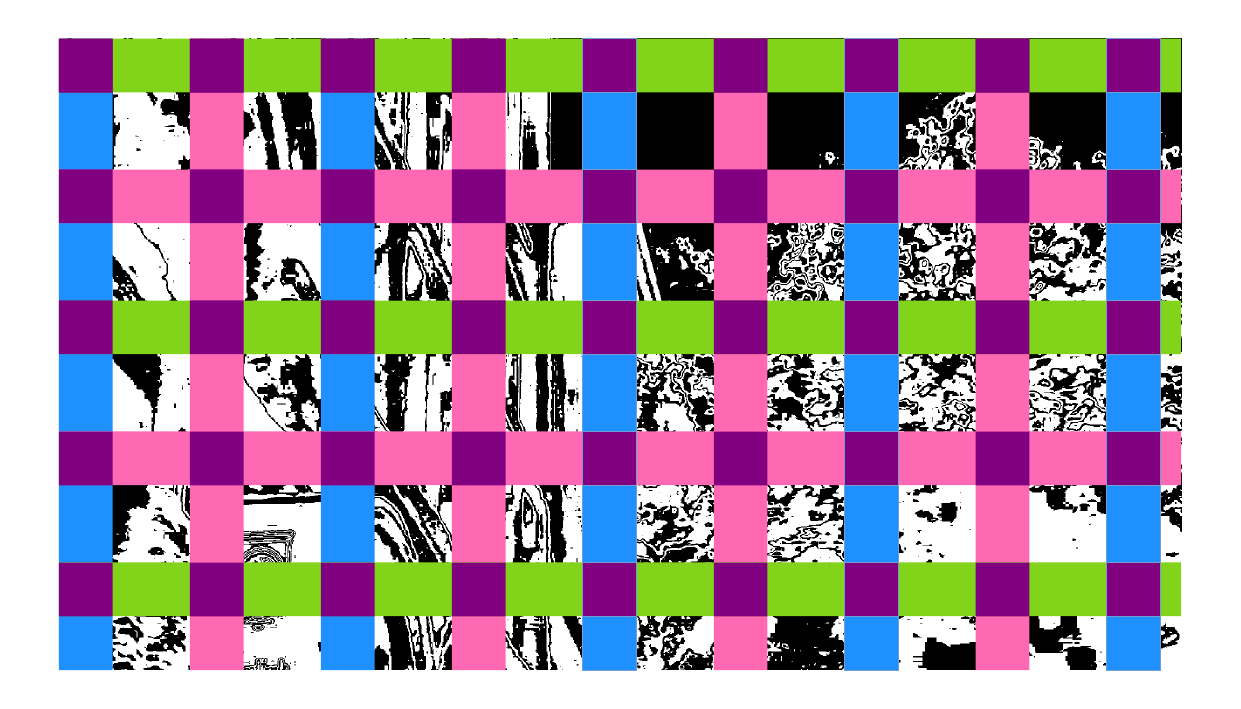} 
  \vspace*{-0.3cm}
 \caption{Embedded area of the pilot signal:
 Values $-1$, $0$, and $+1$ are embedded in the blue, pink, and green areas as a pilot signal.}
  \label{fig:pilot_signal}
\end{figure}

The original color image is decomposed into its YUV components, and the watermark is embedded into the Y-component, the luminance value, using the QIM~\cite{qim}. 
The pilot signal is embedded into the V- or U-component to avoid degradation of the watermark.
In the following description, the U-component is chosen.
Let the image size be $L_w\times L_h$ pixels and the pixel value of the U-component at the coordinate $(x,y)$ be $U(x,y)$. 
Here, the embedded area of the pilot signal is shown in Fig.\ref{fig:pilot_signal}. The width and interval of each color grid line are 5 and $\gamma=100$ pixels, respectively. 
The proposed pilot signal is a three-valued signal with the values $-1, 0$, and $+1$. 
The vertical and horizontal directions are distinguished by grid lines with different combinations of values. Specifically, the vertical consists of lines with values of $-1$ and 0. The horizontal consists of lines with values of 0 and $+1$.
The blue areas are filled with the value $-1$, the pink areas are filled with the value $0$, and the green areas are filled with the value $+1$. The overlapping areas in purple are filled with alternating values of $-1, 0$, and $+1$.
The pixel values in the other areas remain unchanged.

With the QIM embedder~\cite{qim}, the pilot signal is embedded by quantizing the pixel value with step width $\Delta$. Then, the pilot signal value $p\in\{-1,0,+1\}$ can be embedded in the pixel value $U(x,y)$. Let $U'(x,y)$ be the pixel value after embedding. It is given by 
\begin{eqnarray}
    U'(x,y) &=&
    \Delta\left( \left\lfloor \frac{U(x,y)}{\Delta}-\frac{p+1}{3} +0.5\right\rfloor
    +\frac{p+1}{3} \right), 
\end{eqnarray}
where $\left\lfloor\cdot\right\rfloor$ denotes the floor function.
In this paper, the step size is set to $\Delta=9$.
Since the usual QIM embedder uses a non-negative integer as the embedding value $p$, this method uses $p+1$ as a modified version.

\subsection{Attack estimation method} \label{subsec:attackEstimation}
The extracted signal $\hat{p}(x,y)\in\left\{-1,0,+1\right\}$ of the tri-level by the QIM extractor~\cite{qim} is extracted from a U-component image $U'(x,y)$ of a stego image. The extracted signal $\hat{p}(x,y)$ is given by
\begin{eqnarray}   \label{eq:ext_qim}
  \hat{p}(x,y) &=& \left(\left\lfloor \frac{3U'(x,y)}{\Delta}
  +0.5 \right\rfloor \mathrm{mod}\; 3\right)-1 .
\end{eqnarray}
The process of subtracting 1 is appended to the extractor at the end of the calculation, since the embedder has embedded the value of $p+1$.

\begin{figure}[tb]
    \centering
  \includegraphics[width=0.5\linewidth]{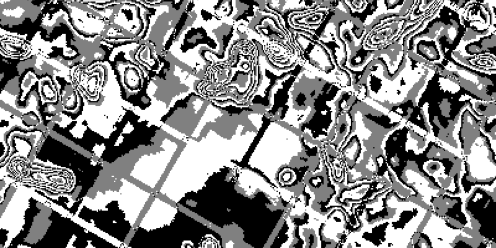}
  \caption{Example of extracted signal: black, gray, and white pixels represent the $-1$, $0$, and $+1$ values extracted by the QIM extractor.}
 \label{fig:extractedSignal}
\end{figure}

Figure~\ref{fig:extractedSignal} shows an example of the extracted signal. The black, gray, and white pixels in the figure represent the $-1$, 0, and $+1$ values extracted by the QIM extractor (\ref{eq:ext_qim}). The signal contains both the grid-shaped pilot signal and the original image component.
Since the grid lines of the pilot signal appear as straight lines, we can see that the geometric transformation can be estimated by detecting them.
In the proposed method, the grid lines with different values are embedded in the vertical and horizontal directions. Specifically, the vertical grid lines have values of $-1$ and $0$, while the horizontal grid lines have values of $0$ and $+1$. Therefore, the extracted signals are divided into two groups to detect the slope and interval of the grid lines. This allows the transformation matrix to be estimated.

\subsubsection{Radon transform} \label{subsubsec:radon}

By Radon transforming the extracted signal $\hat{p}(x,y)$, the location and slope of the grid lines of the pilot signal can be detected~\cite{apsipa2024}.
The Radon coefficient $R(\phi,\rho)$ can be calculated by
\begin{eqnarray} \label{eq:defRadon}
    R(\phi,\rho) &=& \int_{-\infty}^{\infty} 
    \hat{p}\left(\phi\cos\theta-u\sin\theta, \phi\sin\theta+u\cos\theta\right) du,
\end{eqnarray}
where $\rho$ is the projection position and $\phi$ is the projection angle. 
The range of the projection angle of the Radon transform is $0\leq \phi\leq180$ degrees, since the intensity distribution is rotationally symmetric.
In the proposed method, since the pilot signal is embedded in the grid structure, a strong intensity of the Radon coefficients can be obtained by line integration in the direction parallel to the grid orientation.
This property can be used to estimate the rotation angle. 

\begin{figure}[tb]
    \centering
    \includegraphics[width=0.8\linewidth]{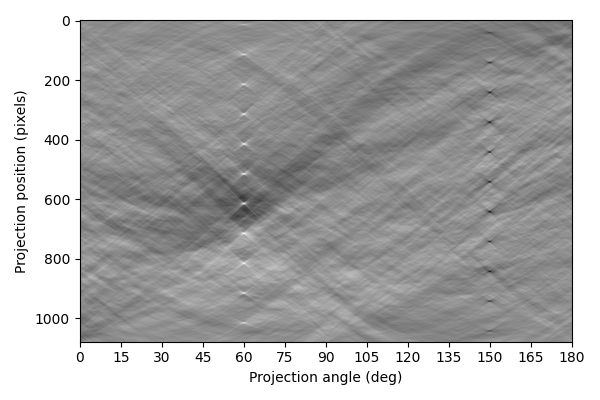}
    \caption{Radon coefficient for the extracted signal: horizontal and vertical axes represent the projection angle and position, respectively.} \label{fig:radon}
\end{figure}

\begin{figure}[tb]
    \centering
    \includegraphics[width=0.6\linewidth]{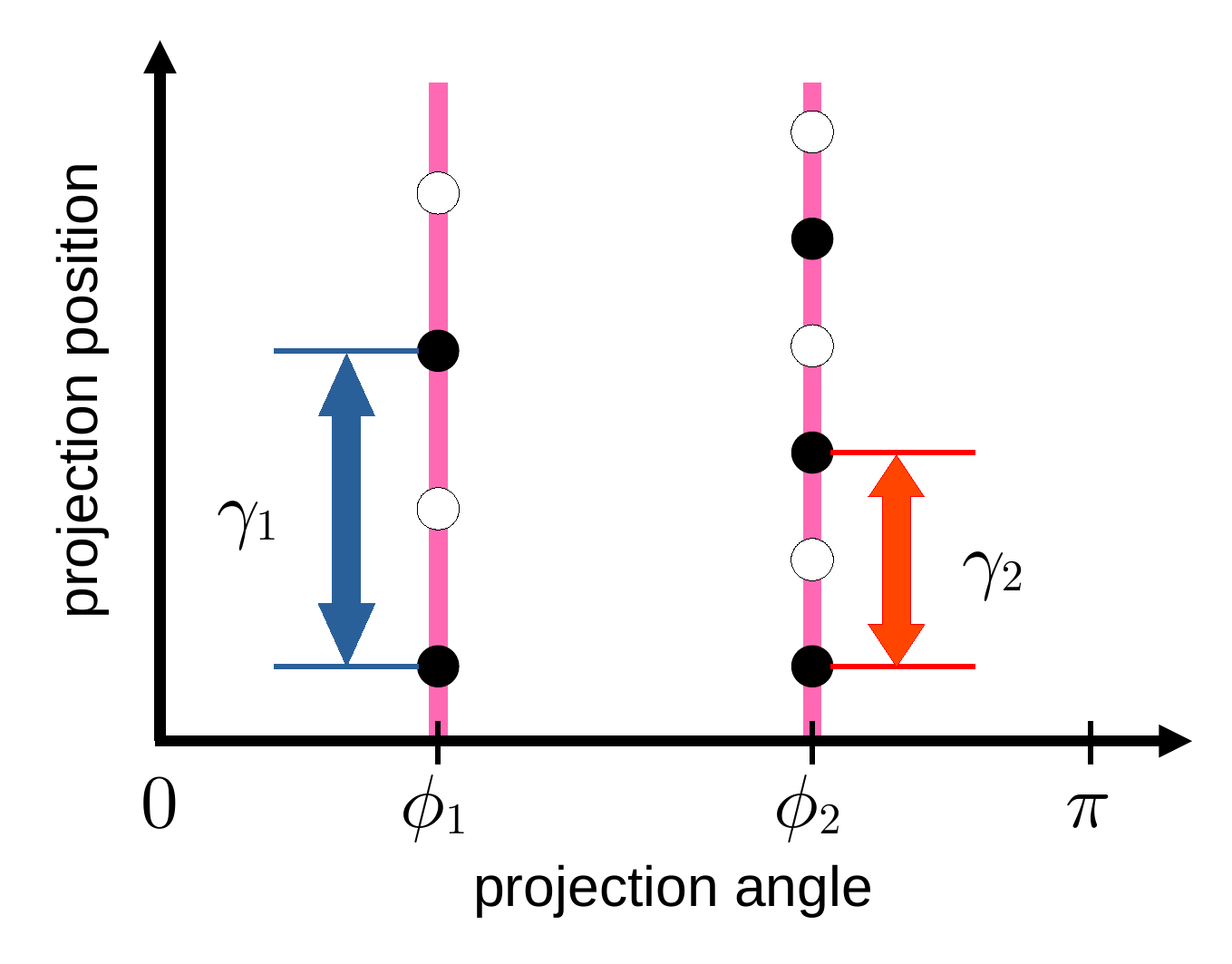}
    \vspace*{-0.3cm}
    \caption{Schematic drawing of Radon coefficient: detection angles and detection intervals are shown.}
\label{fig:radonSchematicDiagram}
\end{figure}

Figure~\ref{fig:radon} shows the Radon coefficient for the extracted signal. The horizontal and vertical axes represent the projection angle and position, respectively.
A series of strong intensity points can be seen at 60 and 150 degrees. A schematic of this is shown in Figure~\ref{fig:radonSchematicDiagram}.
These strong intensities appear at two angles $\phi_1$ and $\phi_2$ $(\phi_1<\phi_2)$.
These angles are called detection angles. They represent the slopes of the grid lines in the two directions. In addition, at each detection angle, strong intensities appear at equal intervals in the direction of the projection position.
We call these intervals the detection intervals, $\gamma_1$ and $\gamma_2$, since they represent the intervals of the grid lines in the two directions.

\subsection{Finding detection angles} \label{subsubsec:detectionAngle}

\subsubsection{Variance of Radon coefficient}
\begin{figure}[tb]
\begin{minipage}[b]{0.5\linewidth}
    \centering
    \includegraphics[width=\linewidth]{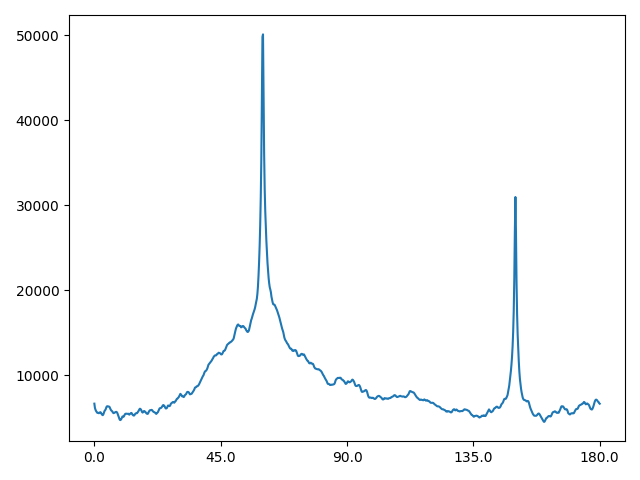}
    \subcaption{Original }
\end{minipage}
\begin{minipage}[b]{0.5\linewidth}
    \centering
    \includegraphics[width=\linewidth]{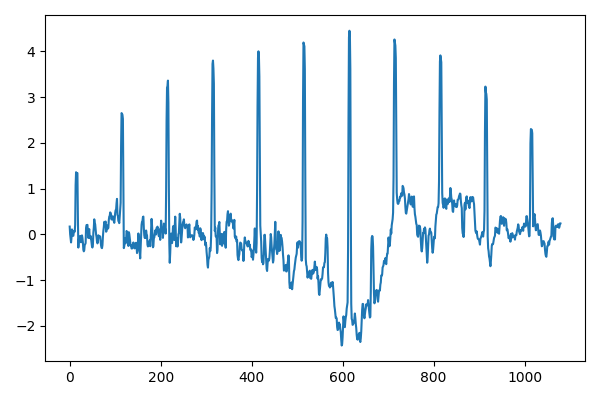}
    \subcaption{Normalized Radon coefficient}
\end{minipage}

\centering
\begin{minipage}[b]{0.5\linewidth}
    \centering
    \includegraphics[width=\linewidth]{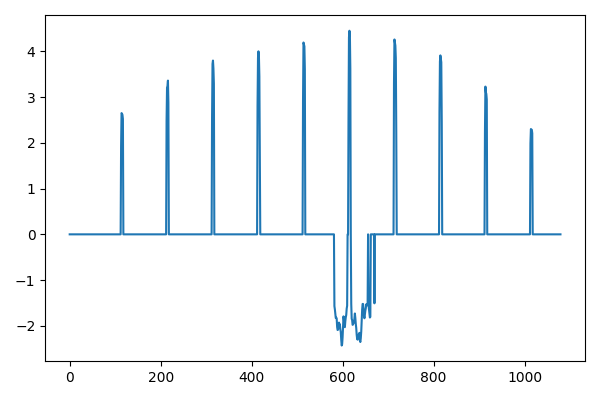}
    \subcaption{Radon coefficients with values below the threshold set to zero}
\end{minipage}
\caption{Variances of Radon coefficients}
\label{fig:executionExample}
\end{figure}

First, we find the detection angles $\phi_1$ and $\phi_2$ from the Radon coefficients.
Focusing the Radon coefficients $R(\phi_1,\rho)$ and $R(\phi_2,\rho)$ on the detection angles, strong intensities appear at equal intervals.
Therefore, the variances of the Radon coefficient are calculated for all projection angles $\phi$, and if the angles with the largest variance are selected, they should be the detection angles $\phi_1$ and $\phi_2$. Then the numerical differentiation of the variances of the Radon coefficient is performed, and the detection angles are found by performing the zero crossing on the derivative values.
Let $V_{\phi_i}=\mathrm{Var}_{\rho}\left[R(\phi_i,\rho)\right]$ be the variance of the Radon coefficient with respect to $\rho$ at the $i$-th projection angle $\phi$.
Figure~\ref{fig:executionExample} (a) shows an example of the variance $V_{\phi_i}$ of the Radon coefficient.
The vertical axis represents the variance, while the horizontal axis represents the projection angle $\phi$. Two prominent peaks are observed at certain projection angles.
Let the detection angle $\phi_1$ or $\phi_2$ be the angle $\phi_i$ that satisfies the following conditions:
\begin{eqnarray}
    dV_{\phi_{i-1}} dV_{\phi_{i+1}} &<& 0,
    \label{eq:zeroCrossing}
\end{eqnarray}
where $dV_{\phi_i}$ is the discrete difference of the variance $V_{\phi_{i-1}}$ and $V_{\phi_{i+1}}$ at the projection angle $\phi_i$, as defined by
\begin{eqnarray}
    dV_{\phi_i} &=& 
    \frac{V_{\phi_{i+1}}-V_{\phi_{i-1}}}{2}.
    \label{eq:gradient}
\end{eqnarray}

\subsubsection{Discrimination of grid lines in two directions} 
\label{subsubsec:discriminationSignals}
One of the obtained detection angles $\phi_1$ and $\phi_2$ represents the detection angle $\phi_v$ for vertical grid lines, and the other represents the detection angle $\phi_h$ for horizontal grid lines. In the proposed method, the grid lines are embedded with different values in the vertical and horizontal directions, so that they can be distinguished. Since the extracted signal $\hat{p}(x,y)$ contains both the vertical and horizontal pilot signals as well as the original image components, it is necessary to distinguish between them. The Radon coefficients at the detection angles $\phi_1$ and $\phi_2$ are normalized. The normalized Radon coefficient $R_n(\phi,\rho)$ is given by
\begin{eqnarray}
    \label{eq:normarizedRadon}
    R_n(\phi,\rho) &=& \frac{R(\phi,\rho)-\mu}{\sigma},
\end{eqnarray}
where $\mu$ and $\sigma$ are the mean and standard deviation of the Radon coefficients. 
Figure~\ref{fig:executionExample} (b) shows the normalized Radon coefficients at the detection angle $\phi$. The horizontal axis represents the projection position.
Since the coefficient values on the pilot signal take on large intensities, coefficient values with small intensities below the threshold can be considered as components derived from the original image. These values should be set to 0.
The Radon coefficient $\widetilde{R}(\phi,\rho)$ without the original image component is given by 
\begin{eqnarray}    \label{eq:denoisedRadon}
    \widetilde{R}(\phi,\rho) = 
    \left\{ 
    \begin{array}{ll}
        0, & \left|R_n(\phi,\rho)\right|\leq 1.5\\
        R_n(\phi,\rho), & \mathrm{otherwise}\\
    \end{array}.
    \right.
\end{eqnarray}

Remember that the vertical grid line $p(x,y)$ consists of values of $-1$ or 0, and the horizontal one consists of values of $0$ or $+1$. In other words, the Radon coefficients at the vertical detection angle $\phi_v$ appear with periodic negative peaks, while the coefficients at the horizontal detection angle $\phi_h$ appear with periodic positive peaks, as shown in (c).
Therefore, the vertical detection angle $\phi_v$ and the horizontal detection angle $\phi_h$ can be determined by comparing the number of occurrences of the positive and negative peaks.

\subsection{Finding detection intervals}
\label{subsubsec:detectionInterval}
Next, the detection intervals $\gamma_v$ and $\gamma_h$ are estimated from the extracted signal $\hat{p}(x,y)$ in (\ref{eq:ext_qim}). 
The extracted signal consists of three values and contains an original image component. Therefore, the extracted signal is divided into the vertical signal $\hat{p}_v(x,y)$ and the horizontal signal $\hat{p}_h(x,y)$, and the detection interval is calculated for each detection direction. The vertical and horizontal signals are given by
\begin{eqnarray}     \label{eq:Vsignal}
    \hat{p}_v(x,y)&=& 
    \left\{ 
    \begin{array}{ll}
        -1, & \hat{p}(x,y) = -1\\
        +1, & \hat{p}(x,y) = 0\\  
        0, & \hat{p}(x,y) = +1\\
    \end{array}, \right. 
\end{eqnarray}
and
\begin{eqnarray}    \label{eq:Hsignal}
    \hat{p}_h(x,y) &=& 
    \left\{ 
    \begin{array}{ll}
        0, & \hat{p}(x,y) = -1\\
        -1, & \hat{p}(x,y) = 0\\
        +1, & \hat{p}(x,y) = +1\\
    \end{array}, \right.
\end{eqnarray}
The vertical grid lines take the values of $-1$ and 0. The value of $+1$ is derived from the original image. Therefore, the component derived from the original image is converted to $\hat{p}_v(x,y)=0$, and the components derived from the grid lines are converted to take positive and negative values.
Similarly, the horizontal grid lines take on values of 0 and $+1$. The value of $-1$ is derived from the original image. Therefore, the component derived from the original image is converted to $\hat{p}_h(x,y)=0$ and the component derived from the grid lines is converted to take positive and negative values. Normalizing the vertical and horizontal signals by $\left(\hat{p}_v(x,y)-\mu_v\right)/\sigma_v$ improves the signal-to-noise ratio because many of the original image-derived components are close to zero, where $\mu_v$ and $\sigma_v$ are the mean and standard deviation of the vertical signal. The same applies for the horizontal signal.

The following is a description of the processing for vertical signals, but the process is the same for horizontal signals.
The converted vertical signal $\hat{p}_v(x,y)$ is Radon transformed by (\ref{eq:defRadon}) and normalized by (\ref{eq:normarizedRadon}). In addition, the original image components are removed by (\ref{eq:denoisedRadon}).
The autocorrelation of the Radon coefficient $\widehat{R}_n(\rho,\phi)$ is calculated. The periodic peaks of the autocorrelation are detected.
The method of estimating the grid interval from autocorrelations was proposed in our previous paper. 
For further details, see Ref.~\cite{IEICE2024}. Here, we provide an overview of the procedure.

A discrete Fourier transform (DFT) is performed on the autocorrelation coefficients. From the peak frequencies of the power spectrum of the DFT coefficients, the base frequency $f_0$ can be calculated, which is the inverse of the grid interval. 
However, the real frequencies are not exactly odd multiples, because they contain noise. Therefore, if there is a frequency $f$ among the detected frequencies that satisfies $0.9nf_0\leq f \leq 1.1nf_0$ for an odd integer $n$, then the frequency $f$ is considered to be an odd multiple of the frequency $f_0$.
Consequently, the grid interval can be estimated by $\gamma_v=\frac{1}{\hat{f}_0}$.

\subsection{Estimating transformation matrix}
\label{subsubsec:matrix}

\begin{figure*}[tb]
\begin{minipage}[b]{0.5\linewidth}
    \centering
    \includegraphics[width=0.5\linewidth]{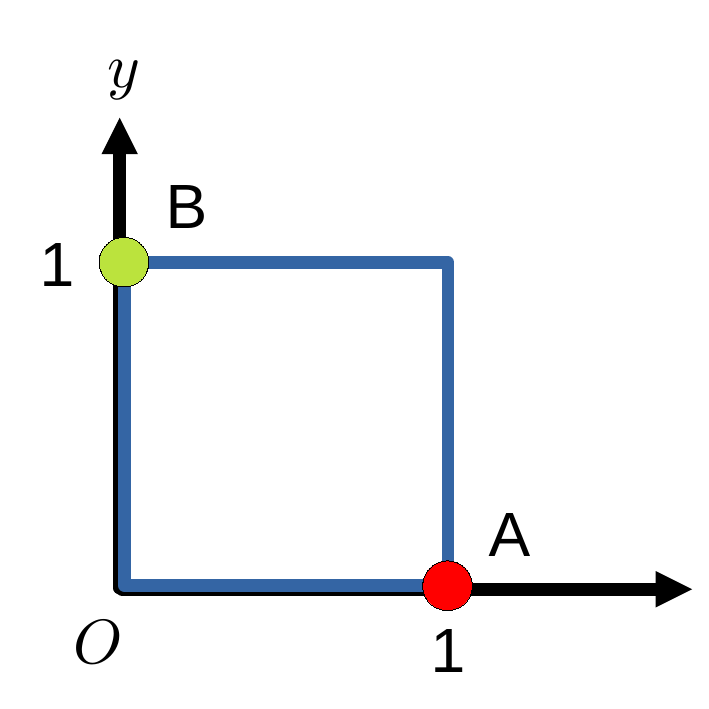}
    \subcaption{Original coordinates}
\end{minipage}
\begin{minipage}[b]{0.5\linewidth}
    \centering
    \includegraphics[width=0.5\linewidth]{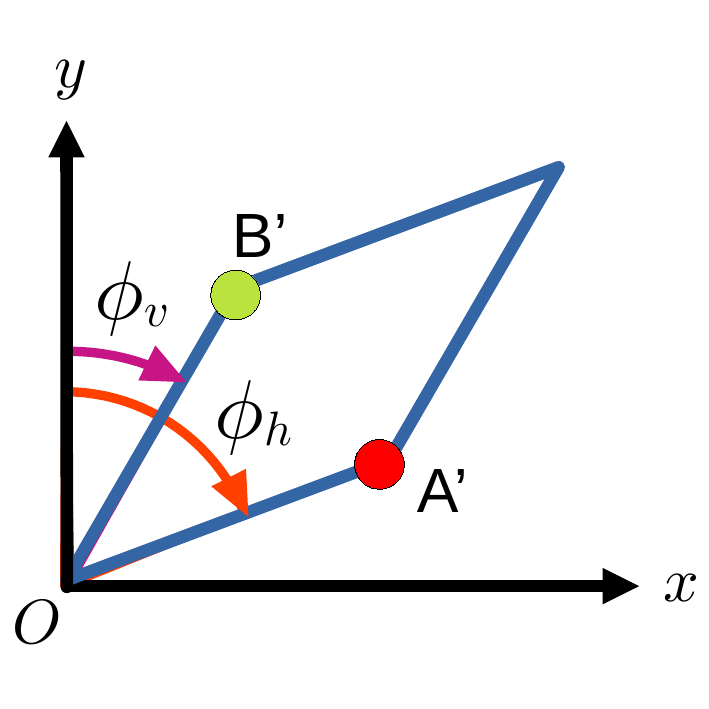}
    \subcaption{Coordinates when the angle is $\phi_v<\phi_h$.}
\end{minipage}

\centering
\begin{minipage}[b]{0.5\linewidth}
    \centering
    \includegraphics[width=0.5\linewidth]{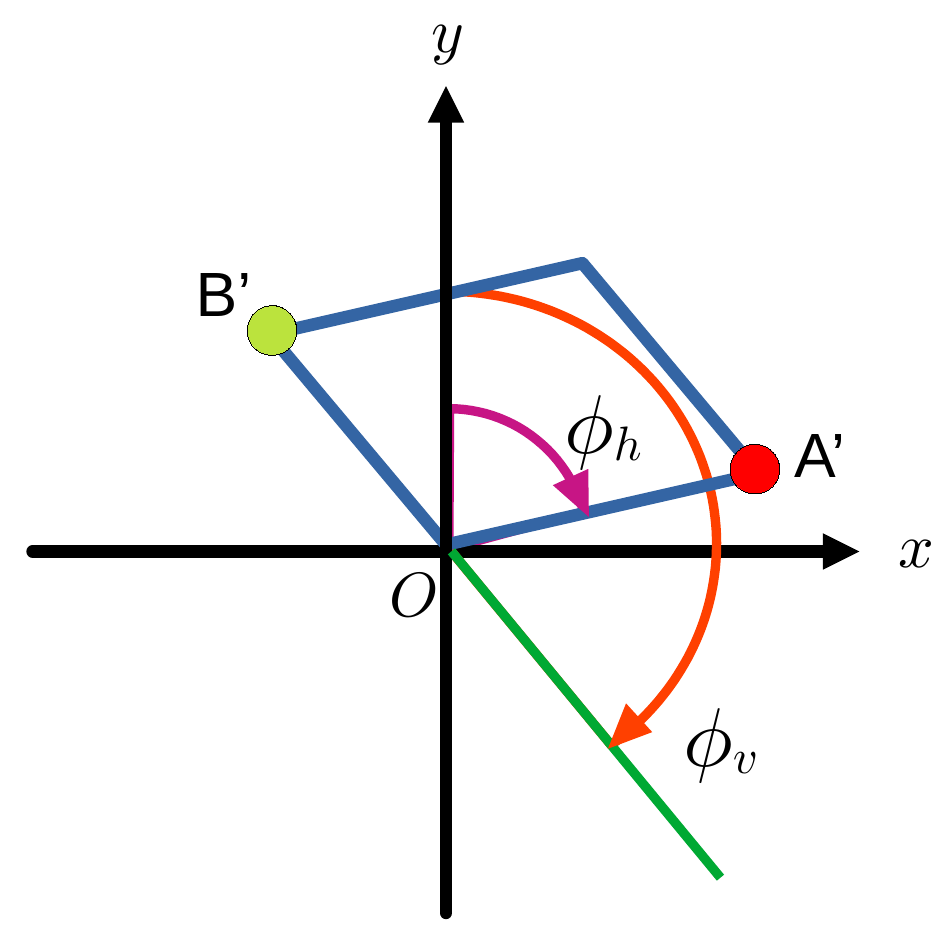}
    \subcaption{Coordinates when the angle is $\phi_h<\phi_v$. }
\end{minipage}
\caption{Coordinate transformation using transformation matrices}
\label{fig:Transform}
\end{figure*}

The next step is to compute the transformation matrix $\bm{T}$ using the detection angles $\phi_v, \phi_h$ and the detection intervals $\gamma_v, \gamma_h$. 
Note that there is a difference between the coordinate system of the Radon coefficients and that of the image.
The transformed detection angles $\phi_v, \phi_h$ are shown in Figure~\ref{fig:Transform}. The direction of rotation associated with the transformation matrix is counter-clockwise, whereas the projection angle in the Radon transform rotates clockwise. It is therefore necessary to take this difference into account when estimating the transformation matrix.
Suppose the coordinates $A(1,0)$ on the horizontal signal and $B(0,1)$ on the vertical signal are transformed using the transformation matrix $\bm{T}$, and the transformed coordinates are $A'(x_a,y_a)$ and $B'(x_b,y_b)$. In this case, the transformation matrix $\bm{T}$ is given by
\begin{eqnarray}
    \bm{T}=\left(
    \begin{array}{ll}
        x_a & x_b\\
        y_a & y_b
    \end{array} \right).
    \label{eq:transformMatrix}
\end{eqnarray}
Let $\alpha$ be the angle between the $x$-axis and the line $OA'$ and $\beta$ be the angle between the $x$-axis and the line $OB'$.
The angles $\alpha$ and $\beta$ correspond to the angles of the horizontal and vertical signals transformed by the matrix $\bm{T}$.

Let us derive the relationship between the detection angles $\phi_v,\phi_h$ and the angles $\alpha, \beta$.
The range of the projection angle of the Radon transform is from 0 to 180 degrees. 
As shown in Figure~\ref{fig:Transform}~(b), if both angles $\alpha$ and $\beta$ are between 0 and 180 degrees, then the detected angle $\phi_v$ is less than the detected angle $\phi_h$. As shown in (c), if the angle $\beta$ is less than 0 degrees, the detected angle $\phi_v$ is greater than the detected angle $\phi_h$. 
Accordingly, we obtain the following equation;
\begin{eqnarray}
(\alpha,\beta) = \left\{
\begin{array}{ll}
    \left(\frac{\pi}{2}-\phi_v, \frac{\pi}{2}-\phi_h\right), & \phi_v<\phi_h\\
    \left(\frac{3\pi}{2}-\phi_v, \frac{\pi}{2}-\phi_h\right), & \mathrm{otherwise}
\end{array}  \right. .
\end{eqnarray}
Note that since the pilot signal is point symmetric, it is not possible to distinguish between the true angle of rotation and an angle rotated by 180 degrees.

\begin{figure}[tb]
    \centering
  \includegraphics[width=0.7\linewidth]{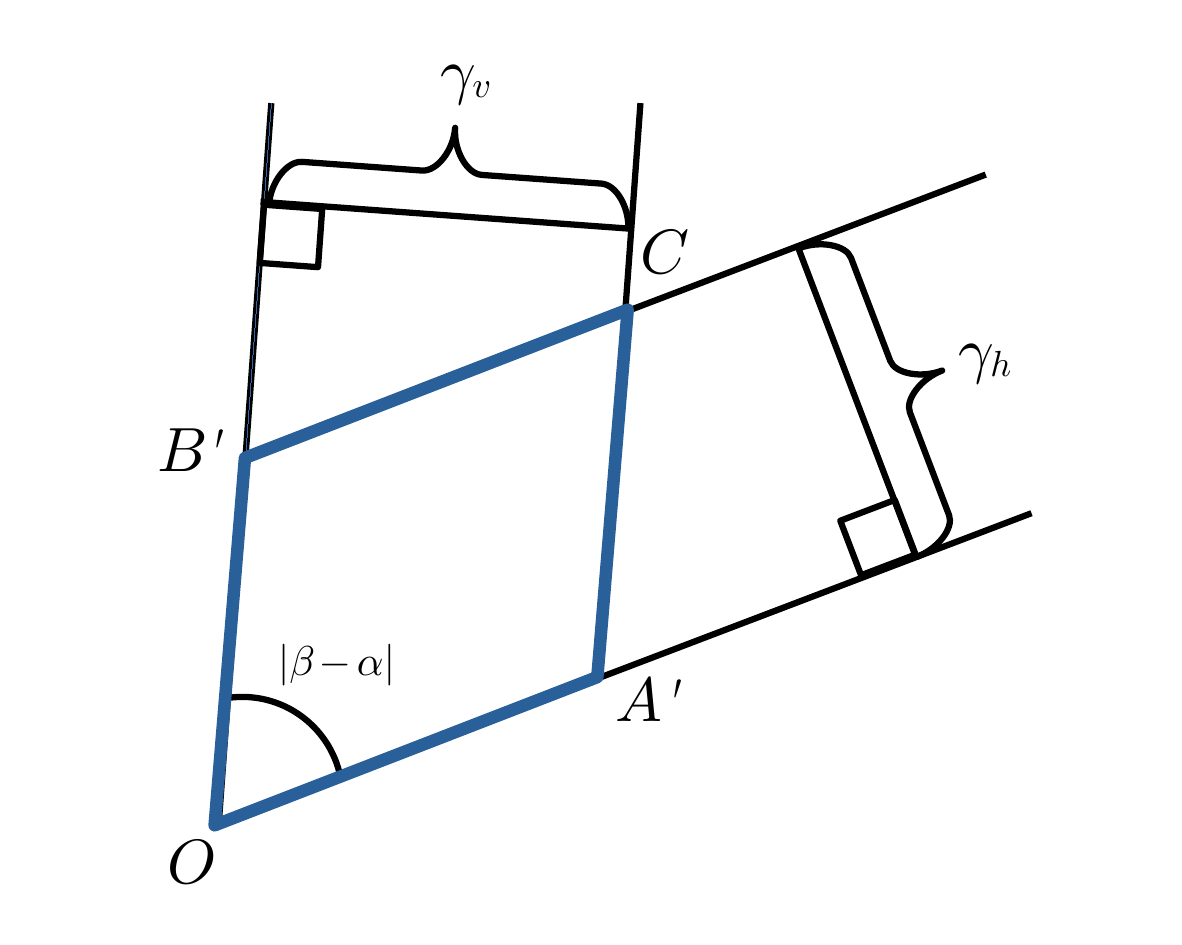}
    \caption{Relationship between the segments $OA'$ and $OB'$, the detection intervals $\gamma_v$ and $\gamma_h$, and the angles $\alpha$ and $\beta$: the lengths of the segments are determined by the detection intervals and angles.}
    \label{fig:pararellogram}
\end{figure}

Next, we calculate the coordinates $A'$ and $B'$ using the detection intervals $\gamma_v,\gamma_h$ and $\alpha,\beta$.
As shown in Figure~\ref{fig:pararellogram}, the distance between the lines $OB'$ and $A'C$ represents the detection interval $\gamma_v$, while the distance between the lines $OA'$ and $B'C$ represents the detection interval $\gamma_h$. Furthermore, the angle between the lines $OA'$ and $OB'$ is given by $|\beta-\alpha|$. Therefore, the lengths of the segments $OA'$ and $OB'$ are given by 
\begin{eqnarray}
    |OA'| = \frac{\gamma_h}{\sin{|\beta-\alpha|}},\  |OB'| = \frac{\gamma_v}{\sin{|\beta-\alpha|}}.
\end{eqnarray}
Accordingly, the coordinates $A'(x_a,y_a)$ and $B'(x_b,y_b)$ are given by 
\begin{eqnarray}
   x_a &=& |OA'|\cos\alpha
           = \frac{\gamma_v\cos\alpha}{\sin|\beta-\alpha|},\\
   y_a &=& |OA'|\sin\alpha
           = \frac{\gamma_v\sin\alpha}{\sin|\beta-\alpha|},
\end{eqnarray}
and
\begin{eqnarray}
   x_b &=& |OB'|\cos\beta
           = \frac{\gamma_h\cos\beta}{\sin|\beta-\alpha|},\\
   y_b &=& |OB'|\sin\beta
           = \frac{\gamma_h\sin\beta}{\sin|\beta-\alpha|}.
\end{eqnarray}
Finally, we obtain the estimated transformation matrix as
\begin{eqnarray}
    \widehat{\bm{T}}= \left(
    \begin{array}{ll}
        \frac{\gamma_v\cos\alpha}{\sin|\beta-\alpha|} & \frac{\gamma_h\cos\beta}{\sin|\beta-\alpha|}\\
        \frac{\gamma_v\sin\alpha}{\sin|\beta-\alpha|} & \frac{\gamma_h\sin\beta}{\sin|\beta-\alpha|}
    \end{array} \right).
    \label{eq:transformMatrix2}
\end{eqnarray}

\section{Evaluation of the proposed method} \label{sec:EstimationSimulation}

In order to validate the correctness of the estimated transformation matrix $\widehat{\bm{T}}$, computer simulations are performed. 
The accuracy is evaluated by the relative error of the estimated matrix given by
\begin{eqnarray}
    E\left(\widehat{\bm{T}};\bm{T}\right) &=& \frac{\| \widehat{\bm{T}}-\bm{T} \|_F}{\| \bm{T} \|_F},
    \label{eq:Terror}
\end{eqnarray}
where $\|\cdot\|_F$ is the Frobenius norm, and that for the $m\times n$ matrix $\bm{T}=\left[t_{ij}\right]$ is defined in
\begin{eqnarray}
    \| \bm{T} \|_F &=& \sqrt{\sum_{i=1}^{m} \sum_{j=1}^{n} |t_{ij}|^2}.    \label{eq:def_frov}
\end{eqnarray}

\subsection{Experimental conditions} \label{subsec:experimental_conditions}
The proposed method is designed for color images. To demonstrate its applicability to photos taken with a smartphone, six high-resolution IHC standard images~\cite{IHC}  ($4608\times 3456$ pixels) were used for evaluation. We also included ten lower-resolution Kodak Lossless True Color images~\cite{Kodak} ($768\times 512$ or $512\times 768$ pixels), which are commonly used in conventional studies for evaluation purposes.
A geometric attack, denoted by the transformation matrix $\bm{T}$, was applied to the stego image.
After the transformation, the center of the image was cropped. The cropping size was $1080\times 1080$ pixels for IHC standard images and $256\times 256$ pixels for Kodak images. 
The pilot signal is designed to be point-symmetric, so the true angle can be estimated by adding 180 degrees. In this case it is also considered as the correct value.
Therefore, after estimating two different transformation matrices, the one with the smaller relative error with respect to the Frobenius norm was used as the estimated matrix.
For each attack, the accuracy of the estimated transformation matrix was evaluated based on its relative error across these cropped images.

\subsection{Grid Interval Determination} \label{subsec:gridInterval}
The grid interval is an important factor that directly affects the performance of the proposed method. The appropriate grid interval depends on the cropping size because geometric transformations change the size of the image, and cropping may result in the loss of grid lines. A narrow grid interval enables the detection of more grid lines after an attack but decreases image quality. Conversely, a wider interval increases the risk of failing to detect enough grid lines. Therefore, the appropriate grid interval must be determined for high- and low-resolution images, considering the accuracy of the estimated transformation matrix and the resulting image quality.

\begin{figure*}[tbp]
\centering
\begin{minipage}[b]{0.8\linewidth}
    \centering
    \includegraphics[width=1\linewidth]{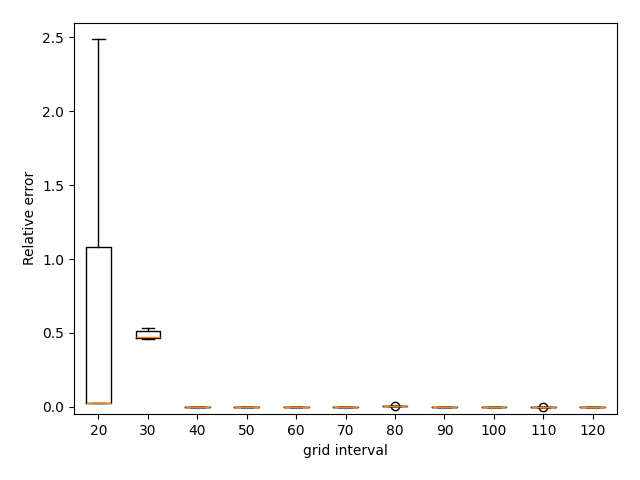}
    \subcaption{Results for high-resolution images}
\end{minipage}

\centering
\begin{minipage}[b]{0.8\linewidth}
    \centering
    \includegraphics[width=1\linewidth]{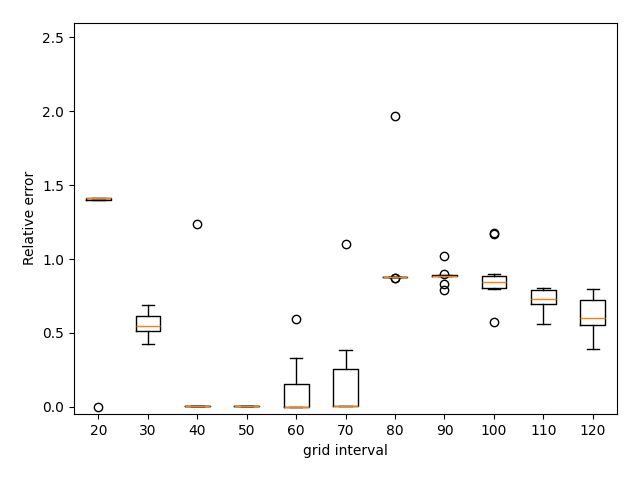}
    \subcaption{Results for low-resolution image}
\end{minipage}
\caption{Accuracy of the estimated transformation matrix for (a) high-resolution and (b) low-resolution images.}
\label{fig:frov-grid}
\end{figure*}

To determine the grid interval, pilot signals with various intervals ranging from 40 to 120 pixels were embedded. Also, the grid line width is 5 pixels. The transformation matrix was estimated from images that were only cropped; no geometric attacks were applied. 
Figure~\ref{fig:frov-grid} shows the relative error of the estimated transformation matrix for different grid intervals in order to determine the appropriate one. (a) shows the results for high-resolution images, and (b) shows the results for low-resolution images.The horizontal axis represents the grid interval, and the vertical axis represents the relative error. The box-and-whisker diagram for each grid interval shows the statistics of the images. The orange lines show the median values. If no pilot signal is detected in the cropped image, it is excluded from the statistics. In low-resolution images, the pilot signal could not be detected in one image with a grid interval of 80 pixels and in two images with a grid interval of 120 pixels. The pilot signal was detectable in the other images.
As a result, the interval could be estimated with a small relative error for high-resolution images with large cropping sizes, even when the grid interval was large. Conversely, for low-resolution images with small cropping sizes, the relative error increased when the grid interval was 80 pixels or more. In both cases, the peak of the Radon coefficient doubled due to the grid width being 5 pixels when the interval was 30 pixels or less. This resulted in a false detection.

\begin{figure}[tbp]
    \centering
    \includegraphics[width=0.8\linewidth]{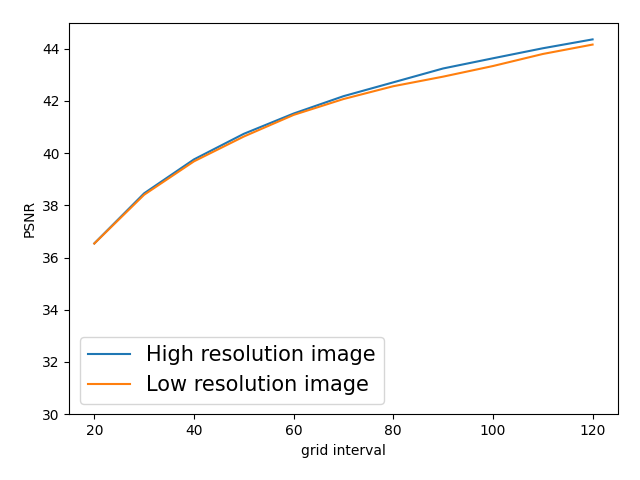}
    \caption{PSNR for different grid intervals in high- and low-resolution images.}
 \label{fig:psnr}
\end{figure}

Figure~\ref{fig:psnr} shows the PSNR versus the grid interval. When the grid interval is narrower, more grid lines are embedded. Thus, image quality degrades.  However, since the area of embedded grid lines per unit area is constant, the image quality stays the same across resolutions.
The image quality should be high, and the estimation matrix should have a sufficiently small relative error. It should also be assumed that the image will be distorted. Therefore, we empirically selected grid intervals of 50 and 100 pixels for low- and high-resolution images, respectively. The average PSNR was $40.6$ dB for the low-resolution images and $43.6$ dB for the high-resolution images.

\subsection{Estimated matrix for single attack} \label{subsec:simpleAttack}

The estimated matrix is evaluated by applying a single attack.
The stego image is subjected to one of the following attacks: anisotropic scaling $(S_x,S_y)$, rotation $\theta_r$, or shearing $\theta_y$. Additionally, it is cropped.
Table~\ref{tab:attacks} shows the parameters of each attack.
For the evaluation of anisotropic scaling, the scaling rate $S_x$ along the $x$-axis is fixed at $S_x=1.0$ and only the scaling rate $S_y$ along the $y$-axis is varied.
Similarly, for symmetry, we consider only $y$-axis shear transformations.

\begin{table}[tb]
    \centering
    \caption{Attack parameters}
    \label{tab:attacks}
    \begin{tabular}{|l|l|}\hline
       Type of attack & Parameter values \\ \hline
       Anisotropic scaling & $S_x=1.0$, $S_y=0,1, 0.2, \cdots 2.0$ \\ \hline
       Rotation & $\theta_r=0, 5, 10, \cdots, 90$ \\ \hline
       Shearing & $\theta_x=0$, $\theta_y=0, 5,10, \cdots, 80$ \\ \hline
     \end{tabular}
\end{table}

The transformation matrix for applying anisotropic scaling is given by 
\begin{eqnarray}
    \bm{T}_m = 
    \left(
    \begin{array}{cc}
        S_x & 0\\
        0 & S_y
    \end{array} \right).
    \label{eq:magnificationMatrix}
\end{eqnarray}

\begin{figure*}[tbp]
\centering
\begin{minipage}[b]{0.8\linewidth}
    \centering
    \includegraphics[width=1\linewidth]{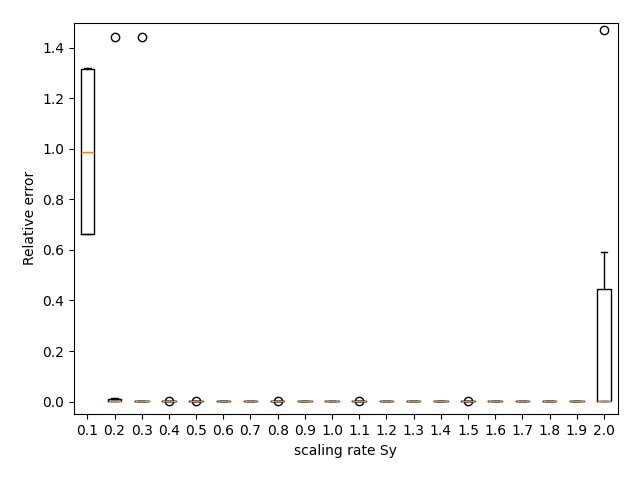}
    \subcaption{Results of high-resolution images}
\end{minipage}

\centering
\begin{minipage}[b]{0.8\linewidth}
    \centering
    \includegraphics[width=1\linewidth]{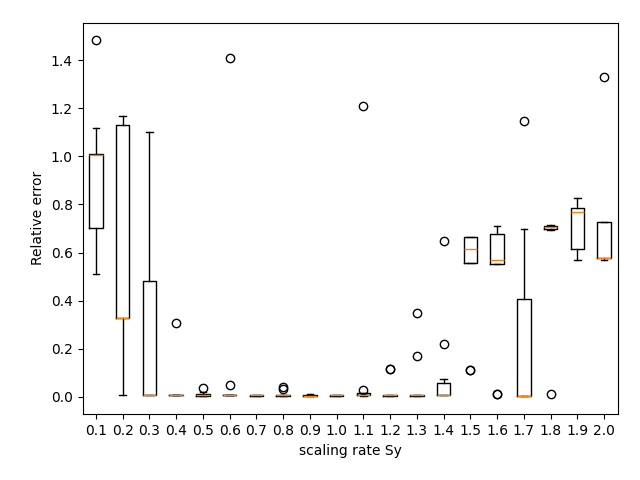}
    \subcaption{Results of low-resolution images}
\end{minipage}
\caption{Anisotropic scaling estimation results: relative error versus scaling rate is shown.}
\label{fig:frov-scaling}
\end{figure*}

The relative error $E\left(\widehat{\bm{T}};\bm{T}_m\right)$ to this matrix is shown in Figure~\ref{fig:frov-scaling}. 
(a) shows the results for high-resolution images and (b) shows the results for low-resolution images.
The horizontal axis represents the scaling rate $S_y$, and the vertical axis represents the relative error.
The box-and-whisker diagram for each scaling rate $S_y$ shows the statistics for the images. However, if the pilot signal could not be detected, the result is excluded from the statistics. The orange lines show the median values. 
The relative error was 0 for almost all of the high-resolution images when the scaling rate $S_y$ was between 0.2 and 1.9.
When the scaling rate was 0.1, the relative error was larger because the detection angles could not be detected in two of the six images. When the scaling rate is small, the reason for failing to detect the pilot signal is the difficulty in estimating the detection interval due to the thinner grid lines.
When the scaling rate is small, the reason for the pilot signal detection failure is the difficulty in estimating the detection interval due to the thinner grid lines.
On the other hand, when the scaling rate is large, the reason for failing to detect the pilot signal is due to the reduced number of grid lines in the cropped image. If the number of grid lines in the cropped image is small, the detection angle and detection interval are likely to be incorrect due to the original image component.
On the other hand, the relative error was nearly zero for many low-resolution images when the scaling rate $S_y$ was between 0.4 and 1.3. However, the matrix could not be estimated for one out of ten images at scaling rates of 0.1, 0.2, 1.9, and 2.0.
Matrix estimation is more difficult for low-resolution images than for high-resolution ones because there are fewer grid lines in the cropped image.

\begin{figure*}[tbp]
\centering
\begin{minipage}[b]{0.8\linewidth}
    \centering
    \includegraphics[width=1\linewidth]{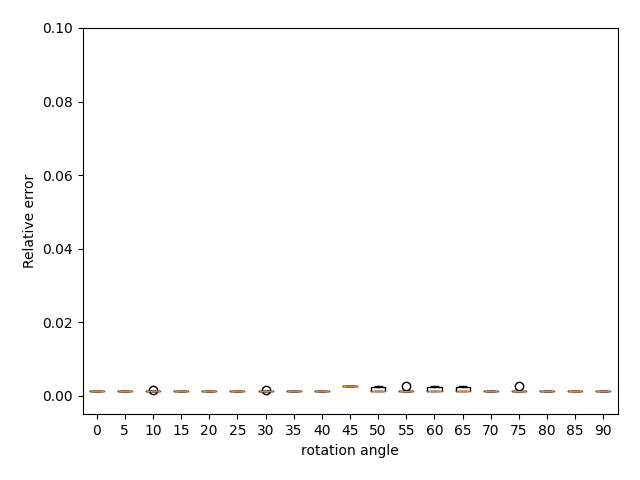}
    \subcaption{Results of high-resolution images}
\end{minipage}

\centering
\begin{minipage}[b]{0.8\linewidth}
    \centering
    \includegraphics[width=1\linewidth]{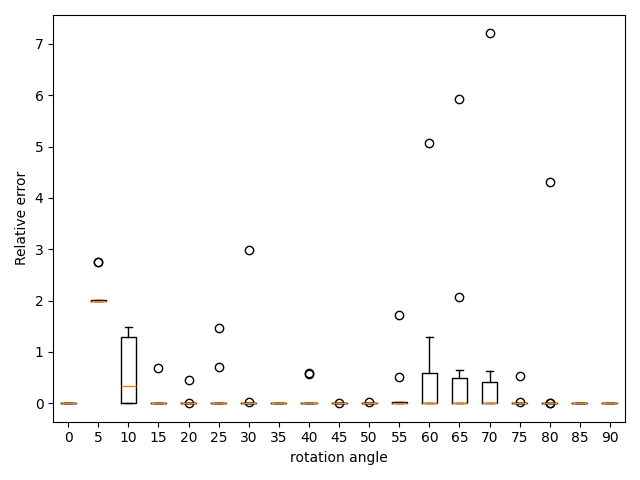}
    \subcaption{Results of low-resolution images}
\end{minipage}
\caption{Rotation angle estimation results: relative error versus rotation angle is shown.}
\label{fig:frov-rotation}
\end{figure*}

Next, the results for the rotational attack are shown.
The transformation matrix of the rotation transformation is given by 
\begin{eqnarray}
    \bm{T}_r = \left(
    \begin{array}{cc}
        \cos \theta_r & -\sin \theta_r\\
        \sin \theta_r & \cos \theta_r
    \end{array}
    \right),    \label{eq:rotationMatrix}
\end{eqnarray}
where $\theta_r$ is the rotation angle.
The relative error for the estimated rotation transformation matrix is shown in Figure~\ref{fig:frov-rotation}.
(a) shows the results for high-resolution images and (b) shows the results for low-resolution images.
The horizontal axis represents the rotation angle $\theta_r$ and the vertical axis represents the relative error $E\left(\widehat{\bm{T}};\bm{T}_r\right)$.
For the rotation angle $\theta_r$ shown in Table~\ref{tab:attacks}, the results for images are shown as box-and-whisker plots.
The orange lines are the median values.
As a result, for the high-resolution images, the relative error was nearly zero for rotation angles ranging from 0 to 90 degrees. Conversely, for low-resolution images, the median relative error was nearly zero for all angles. In other words, at least half of the angles could be estimated correctly. However, the number of cases with large estimation errors increased due to the small cropping size.

\begin{figure*}[tbp]
\centering
\begin{minipage}[b]{0.8\linewidth}
    \centering
    \includegraphics[width=1\linewidth]{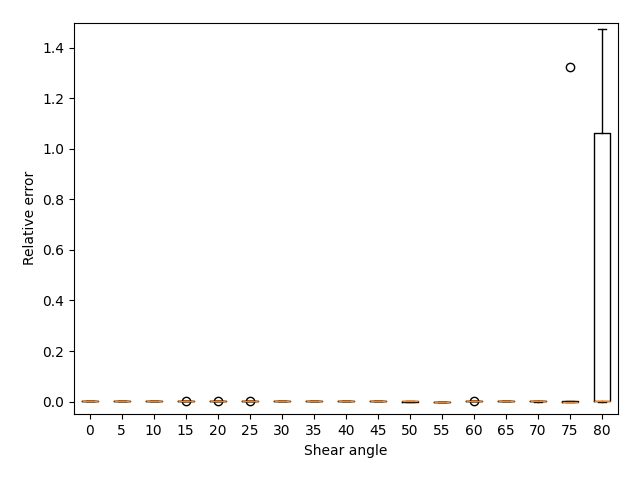}
    \subcaption{Results of high-resolution images}
\end{minipage}

\centering
\begin{minipage}[b]{0.8\linewidth}
    \centering
    \includegraphics[width=1\linewidth]{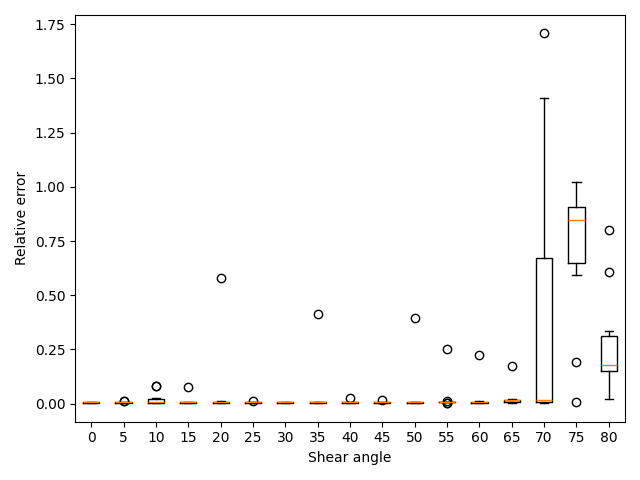}
    \subcaption{Results of low-resolution images}
\end{minipage}
\caption{Shearing angle estimation results: relative error versus shearing angle is shown.}
\label{fig:frov-shearing}
\end{figure*}

Next, the results for the shear attack are shown.
The transformation matrix of the shear transformation in the $y$-direction is given by 
\begin{eqnarray}
    \bm{T}_y = \left(
    \begin{array}{cc}
        1 & 0\\
        \tan{\theta_y} & 1
    \end{array}
    \right),
    \label{eq:shearingMatrix}
\end{eqnarray}
where $\theta_y$ denotes the shear angle in the $y$-direction. 
Similarly, the shear transformation in the $x$-direction is represented by the shear angle $\theta_x$.
The relative error of the shear transformation matrix is shown in Figure~\ref{fig:frov-shearing}.
(a) shows the results for high-resolution images and (b) shows the results for low-resolution images.
The horizontal axis represents the shear angle $\theta_y$ along the $y$ axis, and the vertical axis represents the relative error $E\left(\widehat{\bm{T}};\bm{T}_y\right)$.
The results for images are shown as box-and-whisker plots for the shear angle $\theta_y$ in Table~\ref{tab:attacks}.
As a result, the relative error was nearly zero for high-resolution images at shear angles below 70 degrees, and nearly zero for low-resolution images at angles below 65 degrees.

\subsection{Estimated matrix for composite attacks} \label{subsec:combinedAttack}

We evaluate the estimated transformation matrix for a composite attack. Degraded stego images were generated by applying the composite attack of scaling, rotation, and shearing. After geometric transformation, a cropping attack was also applied to the image. The proposed method was applied to the cropped image and the transformation matrix was estimated by extracting the pilot signal.
The first attack is represented by a transformation matrix $T_1$. Similarly, the second and third attacks are represented by matrices $T_2$ and $T_3$, respectively.
The combined attack is given by the matrix $T_c=T_3T_2T_1$.
The combinations of transformation matrices $T_1$, $T_2$, and $T_3$ are shown in Table~\ref{tab:combinedAttacks}.
Note that in the composite attack, the results vary depending on the order of the transformations. Since there are many possible combinations of attacks, we performed the twelve composite attack patterns listed in Table~\ref{tab:combinedAttacks}.
The presence of the parameters $S_x$ and $S_y$ in the table indicates that a scaling attack was applied. Similarly, the presence of the parameter $\theta_r$ indicates a rotation attack. If the parameters $\theta_x$ or $\theta_y$ appear, they indicate that a shear transformation was applied.
In the example of Pattern 1, the stego image is first transformed by a shear transformation in the $y$-direction, represented by the transformation matrix $T_1$. It is then transformed by a scaling transformation represented by $T_2$, and finally by a shear transformation in the $x$-direction represented by $T_3$. These matrices are given by
\begin{eqnarray}
    T_1=\begin{pmatrix}
        1.0 & 0.0 \\
        \tan\frac{50\pi}{180} & 1.0
    \end{pmatrix},
    T_2=\begin{pmatrix}
        0.6 & 0.0 \\
        0.0 & 1.1
    \end{pmatrix},
    T_3=\begin{pmatrix}
        1.0 & \tan\frac{65\pi}{180} \\
        0.0 & 1.0
    \end{pmatrix}. \label{eq:T1-3}
\end{eqnarray}

\begin{figure*}[tbp]
\centering
\begin{subfigure}[b]{0.3\textwidth} 
\centering
 \includegraphics[width=27mm]{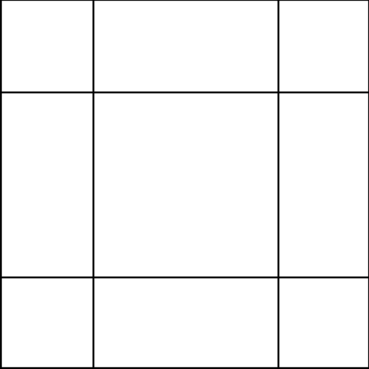}
 \subcaption{Original image\\\mbox{}}
\end{subfigure}
\hfill
\begin{subfigure}[b]{0.3\textwidth} 
\centering
        \includegraphics[width=27mm]{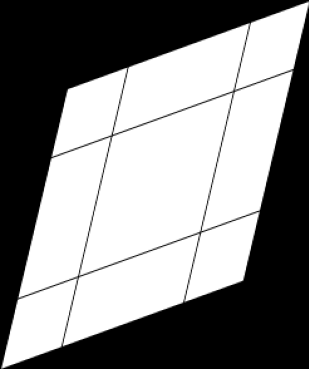}
    \subcaption{Image transformed using Pattern 6}
\end{subfigure}
\hfill
\begin{subfigure}[b]{0.3\textwidth} 
\centering
        \includegraphics[width=27mm]{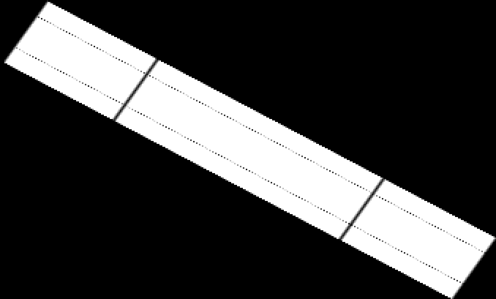}
    \subcaption{Image transformed using Pattern 11}
\end{subfigure}
\caption{Examples of successful and unsuccessful estimation of combined attacks.}
\label{fig:TransformExample}
\end{figure*}

Figure~\ref{fig:TransformExample} shows examples of these transformations. An image with a grid is shown in (a), the result of applying transformation Pattern 6 to this image in (b), and the result of applying transformation Pattern 11 in (c). The image transformation may result in thinner grid lines, which may make it more difficult to detect the lines.

\begin{table}[tbp]
    \centering
    \caption{Combination of attacks: Three attacks were applied in the order $T_1, T_2, T_3$.}  \label{tab:combinedAttacks}
    \begin{tabular}{|l|l|l|l|l|}\hline
         & pattern 1 & pattern 2 & pattern 3  \\\hline
         $T_1$   & $\theta_y=50$ & $\theta_r=30$ &$\theta_y=40$    \\\hline
         $T_2$   & $S_x=0.6,S_y=1.1$ 
        & $\theta_x=30$ &$S_x=1.5,S_y=1.3$    \\\hline
         $T_3$   & $\theta_x=65$ &$\theta_y=65$& $\theta_x=25$  \\\hline\hline
         & pattern 4 & pattern 5 & pattern 6  \\\hline
         $T_1$   & $\theta_x=60$ & $S_x=0.8,S_y=1.4$ &$\theta_y=20$    \\\hline
         $T_2$   & $S_x=1.3,S_y=0.5$ & $\theta_r=250$ &$\theta_x=55$    \\\hline
         $T_3$   & $\theta_y=70$ &$\theta_y=15$& $\theta_r=355$  \\\hline\hline
         & pattern 7 & pattern 8 & pattern 9  \\\hline
         $T_1$   & $\theta_y=70$ & $S_x=1.9,S_y=1.5$ & $S_x=0.7,S_y=1.1$    \\\hline
         $T_2$   & $\theta_r=195$ & $\theta_x=45$ &$\theta_r=215$    \\\hline
         $T_3$   & $S_x=1.2,S_y=1.5$ & $\theta_y=55$ & $\theta_x=60$  \\\hline\hline
         & pattern 10 & pattern 11 & pattern 12  \\\hline
         $T_1$   & $\theta_y=30$ & $S_x=0.9,S_y=0.5$ &$S_x=0.6,S_y=1.1$    \\\hline
         $T_2$   & $\theta_x=65$ & $\theta_r=245$ &$\theta_y=60$    \\\hline
         $T_3$   & $\theta_r=105$ &$\theta_x=20$& $\theta_y=0$  \\\hline
     \end{tabular}
\end{table}

\begin{figure*}[tbp]
\centering
\begin{minipage}[b]{0.8\linewidth}
    \centering
    \includegraphics[width=1\linewidth]{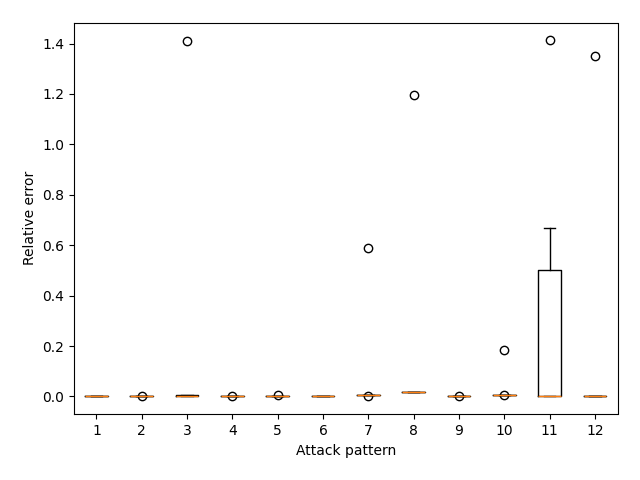}
    \subcaption{Results of high-resolution images}
\end{minipage}

\centering
\begin{minipage}[b]{0.8\linewidth}
    \centering
    \includegraphics[width=1\linewidth]{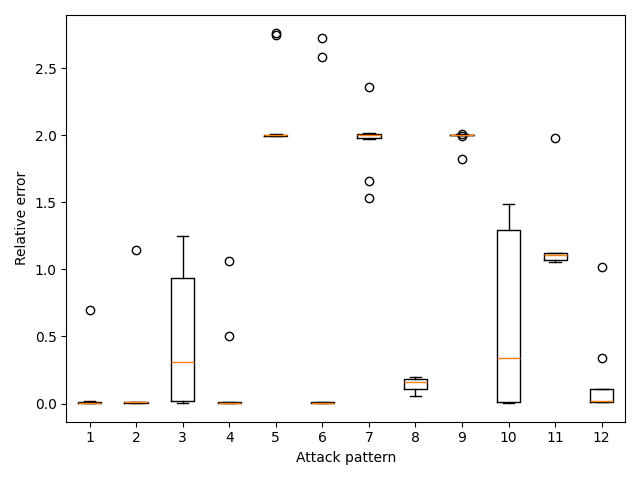}
    \subcaption{Results of low-resolution images}
\end{minipage}
\caption{Estimation results for random attacks: the relative error of the estimation results for each attack is shown.}
\label{fig:frov-random}
\end{figure*}

The relative errors of the estimated transformation matrices for various composite attacks are shown in Figure~\ref{fig:frov-random}. 
(a) shows the results for high-resolution images and (b) shows the results for low-resolution images.
The horizontal axis of the figure represents the number of attack patterns, and the vertical axis represents the relative error.
The results for each attack are shown as box-and-whisker plots. 
The results were generally good for the high-resolution images, including those subjected to strong composite attacks, as shown in Figure~\ref{fig:TransformExample} (c). However, there were many large relative errors for the low-resolution images, though the median relative error was nearly zero. Additionally, the matrix could not be estimated for seven out of ten images for pattern 8, four out of ten for pattern 11, and two out of ten for pattern 12. We find it difficult to estimate the composite attacks in the case of small cropping sizes.

\section{Evaluation of the watermarking method with attack estimation} 
\label{sec:WatermarkSimulation}

\begin{figure}[tb]  \centering 
\begin{minipage}[b]{0.45\linewidth} \centering 
   \includegraphics[width=0.90\linewidth]{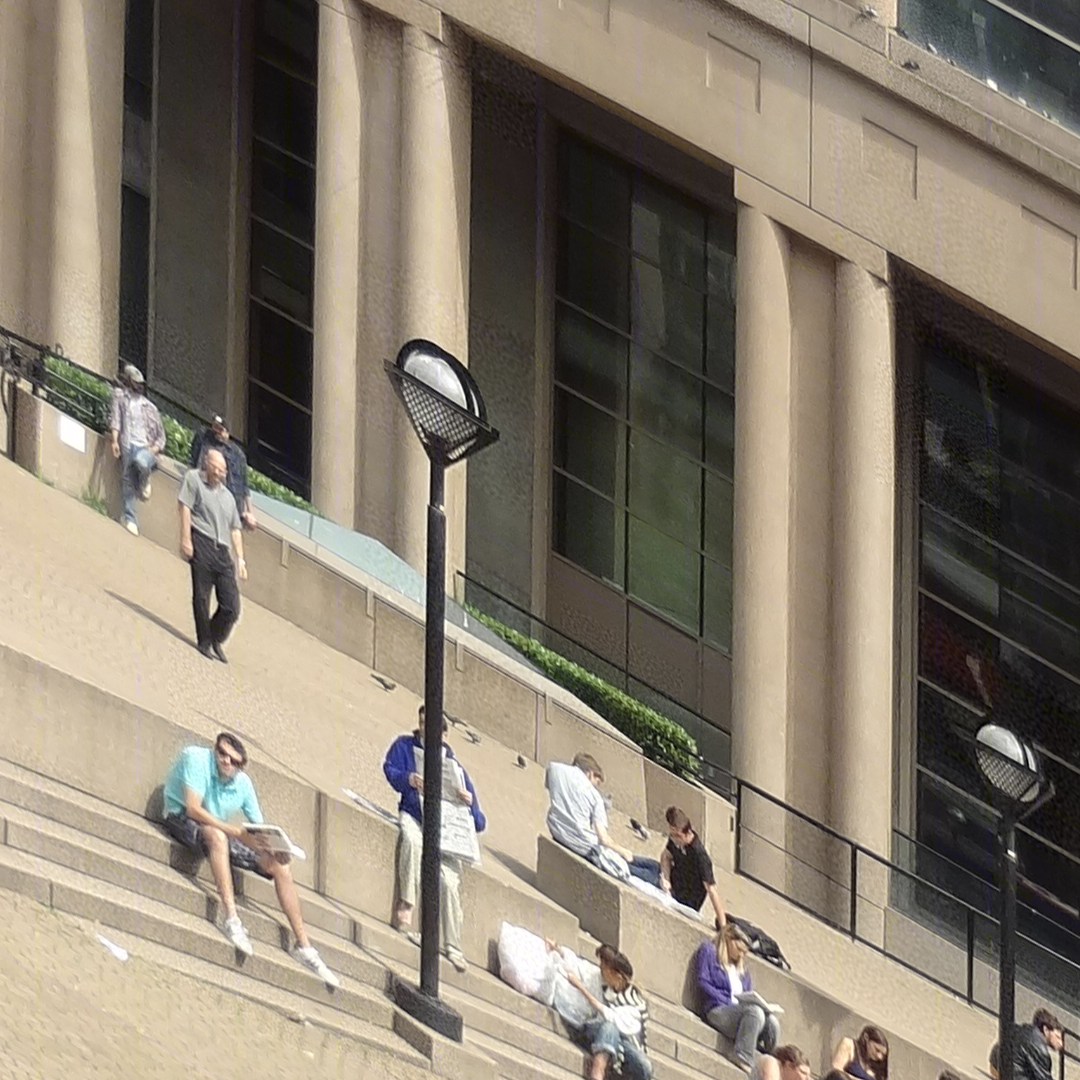}
   \subcaption{Attacked image}
\end{minipage}~\begin{minipage}[b]{0.45\linewidth} \centering 
   \includegraphics[width=0.90\linewidth]{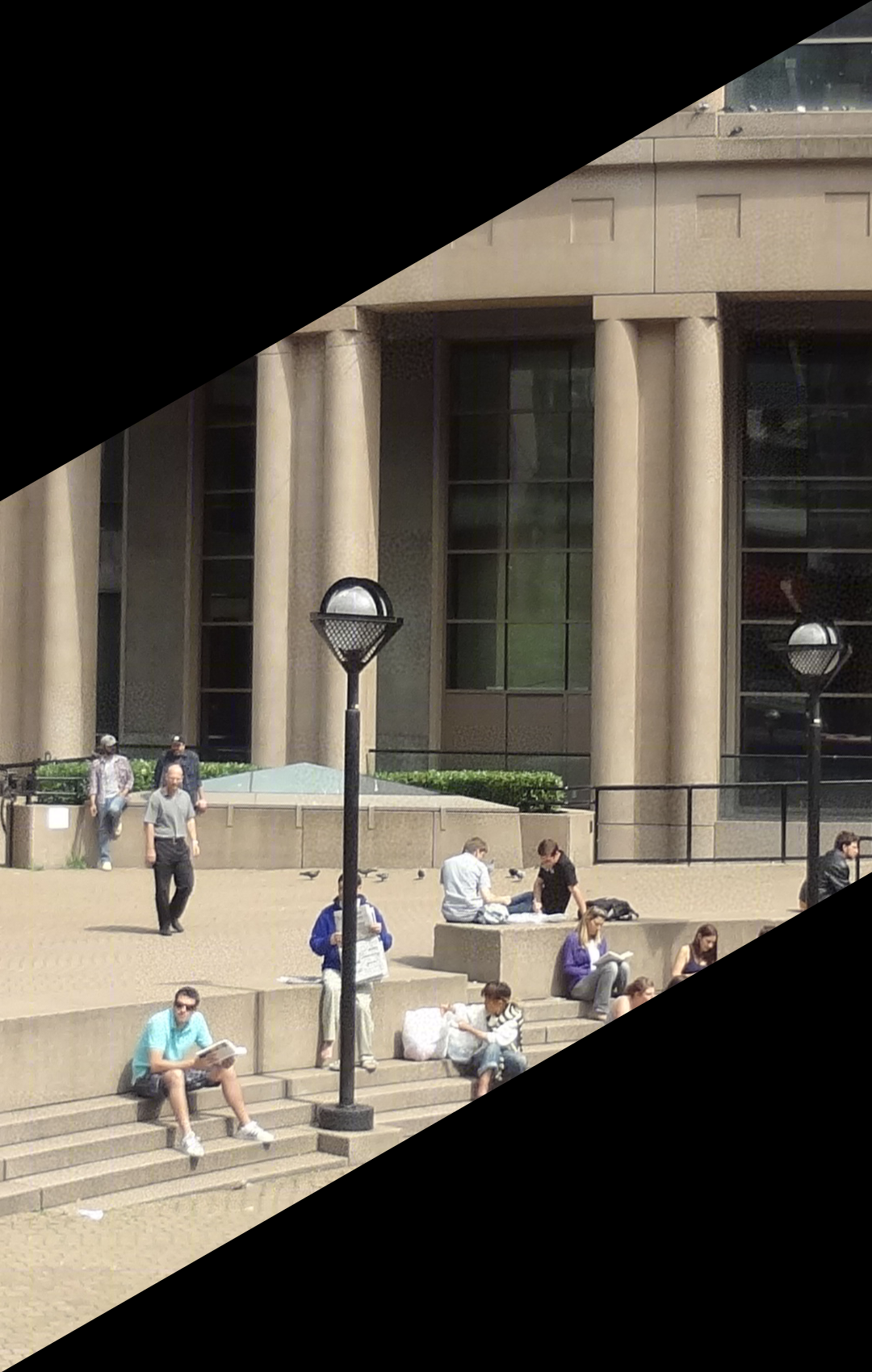}\\
   \subcaption{Inverse transformed image}
\end{minipage}
\caption{Example of an attacked image and its inverse transformed image: This figure shows a degraded image after shearing with an angle of $S_y=30$ degrees and cropping to a size of $1080\times 1080$ pixels. 
Margins resulting from the inverse transformation are filled with black.}
 \label{fig:inverse_image}
\end{figure}

The proposed geometric attack estimation method can be introduced into conventional watermarking methods. In this section, we extract the watermarks from the attacked stego images and evaluate their error rates. The watermarking method that uses SIFT feature points and DFT~\cite{IEICE2024} is considered tolerant of geometric transformations and cropping. In this method, the area around the SIFT feature points of the original image is used as the embedding region. The watermark is then embedded in a ring shape in the coefficients obtained by the DFT of the region. This method has the advantage that the watermark can be extracted even if the image is cropped. However, there is a problem with the conventional method: watermarks are difficult to extract when the shape of the embedding region is distorted by shearing or other geometric attacks. To address this issue, we introduce our attack estimation method. First, the transformation matrix is estimated from the attacked image, as shown in Fig.~\ref{fig:inverse_image} (a). Next, the image is inverted to correct the distortion, as shown in Fig.~\ref{fig:inverse_image} (b). Finally, the watermark is extracted from the inverse transformed image. We evaluate the bit error rate (BER) of the estimated watermarks. We also quantitatively evaluate the image quality using PSNR when both the watermarks and the pilot signal are embedded.

This watermarking method~\cite{IEICE2024} has a limitation. The watermark is embedded in a region ranging from 120 to 180 pixels on each side around the SIFT feature point. To extract the watermark, two or more watermarked regions must be included. Therefore, the cropping size must be large enough. Thus, this section only applies the method to high-resolution images.

\subsection{Experimental conditions} \label{subsec:experimental_conditions2}

Six IHC standard images~\cite{IHC} ($4608\times 3456$ pixels) are used as the original images, and the pilot signal is embedded in these U-components to generate stego images. 
As with the method~\cite{IEICE2024}, a 300-bit watermark is embedded in the Y-components. However, no error-correcting code is used for the watermark.
The stego images were attacked by geometric attacks with transformation matrices $\bm{T}$, and then they were cropped to $1080\times 1080$ pixels from an arbitrary location. We evaluated the average BER of the watermarks extracted from the six cropped images for each attack.

Two transformation matrices are estimated from a single degraded image. These matrices are related in that they are matched by a 180-degree rotation. Since it is not possible to distinguish between them directly, inverse transformed images are generated using each matrix. The matrix with the lower BER in the resulting estimated watermark is adopted as the correct result.
Currently, it is not possible to determine which estimated watermark is correct. This problem can be solved by embedding additional known check bits separately from the watermark~\cite{hayashi}.

\subsection{Image quality evaluation}
First, the image quality is evaluated in terms of PSNR. The average PSNR of the stego images with only a watermark embedded in the original image was $38.29$ dB. In contrast, the average PSNR of the stego images with both the watermark and pilot signal embedded was $37.25$ dB. Therefore, the average degradation in image quality due to the pilot signal was $1.04$ dB.

\subsection{Evaluation against single attacks}

Next, a single attack is applied to a stego image that is embedded with both a pilot signal and a watermark. Then, the watermark is extracted from the attacked image.
The stego image is subjected to one of the following attacks: anisotropic scaling $(S_x,S_y)$, rotation $\theta_r$, or shearing $\theta_y$. Additionally, it is cropped.
Table~\ref{tab:attacks2} shows the parameters of each attack.
For the evaluation of anisotropic scaling, the scaling rate $S_x$ along the $x$-axis is fixed at $S_x=1.0$ and only the scaling rate $S_y$ along the $y$-axis is varied.
Because of symmetry, we consider only $y$-axis shear transformations.

\begin{table}[tb] \centering
    \caption{Attack parameters}
    \label{tab:attacks2}
    \begin{tabular}{|l|l|}\hline
       Type of attack & Parameter values \\ \hline
       Anisotropic scaling & $S_x=1.0$, $S_y=0.5,0.8,1.0,1.3,1.50$ \\ \hline
       Rotation & $\theta_r=0, 30,45,60,90$ \\ \hline
       Shearing & $\theta_x=0$, $\theta_y=5,20,35,50,65$ \\\hline
    \end{tabular}
\end{table}

\begin{figure}[tbp] 
    \centering 
    \includegraphics[width=0.8\linewidth]{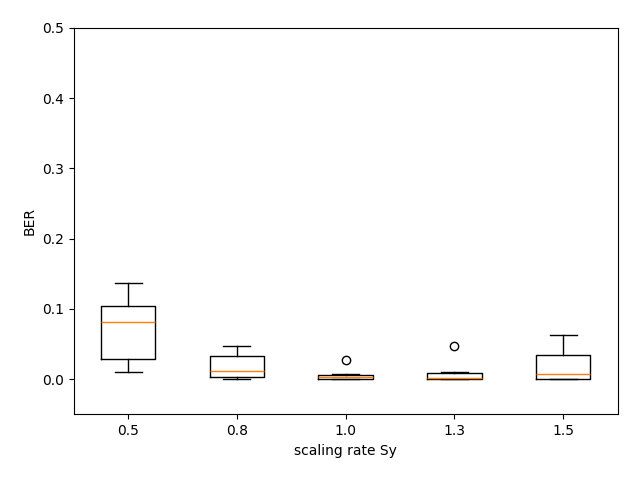}
    \caption{BER of the extracted watermarks against anisotropic scaling attacks}
 \label{fig:BER-scaling}
\end{figure}

Figure~\ref{fig:BER-scaling} shows the BER of the watermarks against the anisotropic scaling attack.
The horizontal axis represents the scaling rate $S_y$, and the vertical axis represents the BER.
The box-and-whisker diagram shows the statistics for the six stego images at each scaling rate $S_y$. The orange lines show the median values. 
As illustrated in the figure, the BER tends to increase as the scaling factor deviates from 1.0. However, the median BER remained consistently below 0.1, indicating that the watermark could be extracted with high accuracy.

\begin{figure}[tbp] \centering
    \includegraphics[width=0.8\linewidth]{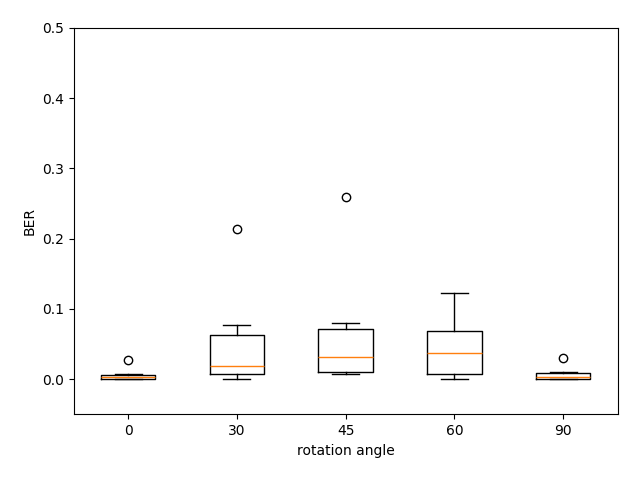}
    \caption{BER of the extracted watermarks against rotation attacks}
 \label{fig:BER-rotation}
\end{figure}

Figure~\ref{fig:BER-rotation} shows the BER of the watermarks against the rotation attack.
The horizontal axis represents the rotation angle $\theta_r$, and the vertical axis represents the BER.
The box-and-whisker diagram shows the statistics for the six stego images at each rotation angle. The orange lines show the median values. 
As shown in the figure, the BER was nearly zero at 0 and 90-degree rotation angles, while it tended to increase at 45 degrees. However, the median BER remained below 0.1 for all angles, indicating that the watermark was extracted with sufficient accuracy.

\begin{figure}[tbp]    
    \centering
    \includegraphics[width=0.8\linewidth]{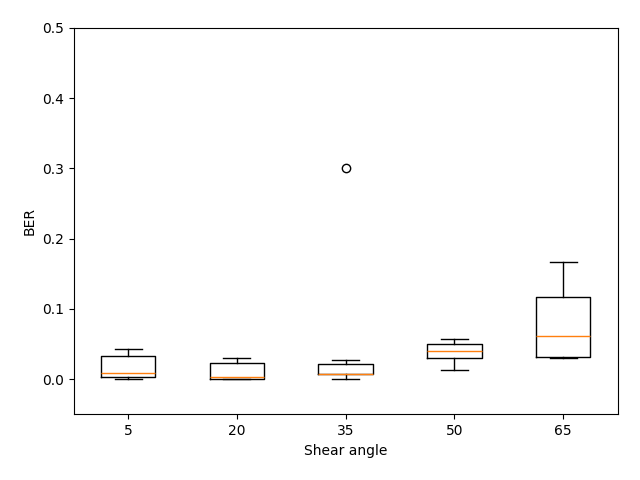}
    \caption{BER of the extracted watermarks against shearing attacks}
 \label{fig:BER-shearing}
\end{figure}

Figure~\ref{fig:BER-shearing} shows the BER of the watermarks against shearing attacks.
The horizontal axis represents the shear angle $\theta_y$, and the vertical axis represents the BER.
As shown in the figure, the median BER was nearly zero for shear angles from 0 to 35 degrees. Conversely, the BER increased as the shear angle increased.

\subsection{Evaluation against composite attacks}

\begin{figure}[tbp]
    \centering
    \includegraphics[width=0.8\linewidth]{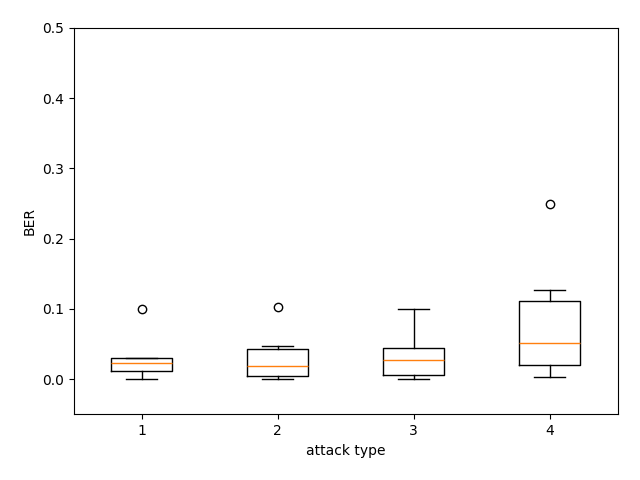}
    \caption{BER of watermark against composite attacks}
    \label{fig:BER-attacks}
\end{figure}

Next, we evaluated performance against composite attacks using BER. Two geometric transformations were applied to the stego image, followed by cropping. Table~\ref{tab:combinedAttacks2} shows the combination of geometric transformations $T_1$ and $T_2$. Figure~\ref{fig:BER-attacks} shows the BER of the watermark for each attack pattern. The horizontal axis represents the attack pattern number and the vertical axis represents the BER. The watermark could be estimated with low BERs for all composite attacks. For example, empirically, as the shear angle increases, the attacked image becomes more distorted, making watermark extraction more difficult. However, the image loses its value when such an attack is applied.

\begin{table}[tb]
    \centering
    \caption{Parameters of composite attacks: Two geometric attacks, $T_1$ and $T_2$, are applied in turn, followed by cropping.}  \label{tab:combinedAttacks2}
    \begin{tabular}{|l|l|l|l|}\hline
              & pattern 1         & pattern 2         \\\hline
        $T_1$ & $\theta_y=5$      & $S_x=1.0,S_y=1.1$ \\\hline
        $T_2$ & $\theta_r=10$     & $\theta_r=10$     \\\hline\hline
              & pattern 3         & pattern 4         \\\hline
        $T_1$ & $\theta_r=5$      & $\theta_x=10$     \\\hline
        $T_2$ & $S_x=1.0,S_y=1.2$ & $\theta_y=5$      \\\hline
     \end{tabular}
\end{table}

\section{Conclusion} \label{sec:conclusion}

In blind watermarking methods, the application of geometric transformations and cropping attacks to the stego image can cause synchronization issues, making it difficult to detect the embedded watermark region. While conventional watermarking methods are considered robust against geometric distortions such as clipping, they often lack robustness against cropping, making it difficult to detect the embedded watermark. Cropping is an unavoidable issue.
Therefore, in this paper, we proposed a method that enabled synchronization of the embedded regions even when the stego image was subjected to geometric transformations and cropping. The proposed method embedded a grid-shaped pilot signal into the image and estimated the transformation matrix based on the distortion of this signal. Furthermore, the estimation accuracy was improved by assigning different values to the vertical and horizontal components of the signal.
The proposed method employs the Radon transform to detect the pilot signal. Given that the signals are grid-shaped, the method leverages the fact that the slope and grid interval of the signals are prominently reflected in the Radon coefficients.

In this paper, we derived a theoretical formula for estimating the geometric transformation matrix based on the detection angle and the interval in the Radon coefficients. The accuracy of this formula was validated through computer simulations.
The proposed method is designed for color images. We evaluated it on six high-resolution images assumed to be smartphone photos and ten low-resolution images commonly used in conventional studies. The grid interval is a very important parameter because geometric transformations change the size of the image, and cropping can result in the loss of grid lines. Thus, appropriate grid intervals were determined for high- and low-resolution images based on estimation matrix accuracy and image quality. As a result, the interval was set to 100 pixels for the high-resolution image and 50 pixels for the low-resolution image.

Geometric transformations including scaling, rotation, and shearing were considered. The accuracy of the estimated matrix was evaluated for both single attacks, where a single geometric transformation and cropping were applied, and composite attacks, where three geometric transformations and cropping were applied.
For high-resolution images, the estimation for a single attack was able to accurately estimate the transformation matrix to a sufficient degree. For composite attacks, the transformation matrices were accurately estimated for all other strengths, except in cases of strong distortions where the stego image was severely degraded.
The estimation results for low-resolution images had larger relative errors compared to high-resolution images. The main reason for this is the smaller cropping size. Nevertheless, the estimation was correct for approximately half of the images.
In summary, the proposed method is effective for high-resolution images, such as those taken with smartphones. For low-resolution images, however, it is necessary to carefully select the appropriate grid interval according to the expected cropping size.

To demonstrate the effectiveness of our proposed method, we applied it to an existing watermarking method~\cite{IEICE2024} that is robust against geometric transformations and cropping. After applying single and composite attacks, we estimated the transformation matrix from the cropped image. Finally, we performed an inverse transformation using the estimated matrix to extract the watermark. Then, we evaluated the accuracy of the watermark. The results confirm that, when the distorted images have practical value, watermarks can be extracted with BERs of less than 0.1. Conversely, there were instances in which the BER of the watermark increased despite an accurate estimation of the transformation matrix. This occurred when the watermark embedding area was physically collapsed by the attack, resulting in the loss of the watermark itself. In such cases, extracting the watermark becomes challenging, even with conventional methods.

\section*{Acknowledgements}
This work was supported by JSPS KAKENHI Grant Numbers JP20K11973 and JP24K15106, the Telecommunications Advancement Foundation, the Yamaguchi University Fund, and the Cooperative Research Project Program, R04/B09 Research on Multifunctional Multimedia Production, of the RIEC, Tohoku University. A part of the computations were performed using the supercomputer facilities at the Research Institute for Information Technology, Kyushu University.


\end{document}